\documentclass{article}

\usepackage[utf8]{inputenc} 
\usepackage[T1]{fontenc}    
\usepackage{hyperref}       
\usepackage{url}            
\usepackage{booktabs}       
\usepackage{amsfonts}       
\usepackage{nicefrac}       
\usepackage{microtype}      
\usepackage{todonotes}
\usepackage{amsmath}
\usepackage{subfig}
\usepackage[lined,boxed,commentsnumbered]{algorithm2e}

\usepackage{color}

\newcommand{\E}{\mathbb{E}}
\newcommand{\RR}{\mathbb{R}}

\title{Multilevel Monte Carlo for Scalable Bayesian Computations\footnote{Corrected version of the preprint }}

\usepackage[margin=0.75in]{geometry}

\author{Tigran Nagapetyan\\
  Department of Statistics, University of Oxford\\
  \texttt{nagapetyan@stats.ox.ac.uk} \\
 \and
 Lukasz Szpruch\\
 School of Mathematics, University of Edinburgh\\
 \texttt{l.szpruch@ed.ac.uk} \\
 \and
 Sebastian J. Vollmer\\
 Department of Statistics, University of Oxford\\
 \texttt{vollmer@stats.ox.ac.uk} \\
  \and
 Konstantinos Zygalakis\\
 School of Mathematics, University of Edinburgh\\
 \texttt{k.zygalakis@ed.ac.uk} \\
   \and
 Mike Giles\\
 Department of Mathematics, University of Oxford\\
 \texttt{mike.giles@maths.ox.ac.uk} \\
}

\begin{document}

\maketitle

\begin{abstract}
  
Markov chain Monte Carlo (MCMC) algorithms are ubiquitous in Bayesian computations. However, they need to access the full data set in order to evaluate the posterior density at every step of the algorithm.  This results in  a great computational burden in big data applications. In contrast to MCMC methods, Stochastic Gradient MCMC (SGMCMC) algorithms such as the Stochastic Gradient Langevin Dynamics (SGLD) only require access to a batch of the data set at every step.  This drastically improves the computational performance and scales well to large data sets. However, the difficulty with SGMCMC algorithms comes from the sensitivity to its parameters which are notoriously difficult to tune. Moreover, the Root Mean Square Error (RMSE) scales as $\mathcal{O}(c^{-\frac{1}{3}})$  as opposed to standard MCMC $\mathcal{O}(c^{-\frac{1}{2}})$ where $c$ is the computational cost.\\
 We introduce a new class of Multilevel Stochastic Gradient Markov chain Monte Carlo algorithms that are able to mitigate the problem of tuning the step size and more importantly of recovering the $\mathcal{O}(c^{-\frac{1}{2}})$ convergence of standard Markov Chain Monte Carlo methods without the need to introduce Metropolis-Hasting steps. A further advantage of this new class of algorithms is that it can easily  be parallelised over a heterogeneous computer architecture. We illustrate our methodology using Bayesian logistic regression and provide numerical evidence that for a prescribed relative RMSE the computational cost is sublinear in the number of data items.

\end{abstract}
\section{Introduction}
In recent years there has been an increasing interest in methods for Bayesian inference which are scalable to Big Data settings. Contrary to optimisation-based or maximum likelihood settings, where one looks for a  single point estimation of parameters, Bayesian methods attempt to obtain a characterisation of the full posterior distribution over the unknown parameters and latent variables in the model. This approach allows for a better characterisation of the uncertainties  inherent to the learning process as well as providing protection against over fitting.  

One of the most widely used classes of methods for Bayesian posterior inference is Markov Chain Monte Carlo (MCMC).  This class of algorithms mixes slowly in complex, high dimensional-models and scales poorly to large data sets \cite{bardenet2015review}. In order to deal with these issues, a lot of effort has been placed on developing MCMC methods that provide more efficient exploration of the posterior, such as Hamiltonian Monte Carlo (HMC) \cite{DUANE1987216,Neal2010MCMC} and its 
Riemannian manifold variant \cite{MGC11}. 

Stochastic gradient variants of such continuous-dynamic samplers have been shown to scale very well with the size of the data sets, as at each iteration they use data subsamples (also called \emph{minibatches}) rather than the full dataset. Stochastic gradient Langevin dynamics (SGLD) \cite{welling2011bayesian} was the first algorithm of this kind showing that adding the right amount of noise to a standard stochastic gradient optimisation algorithm leads to sampling from the true posterior as the step size is decreased to zero. Since its  introduction, there have been a number of 
articles extending this idea to different samplers \cite{Chen2014,BLS16,NIPS2014_5592}, as well as carefully studying the behaviour of the mean square error (MSE) of the SGLD for decreasing step sizes and for a fixed step size \cite{teh2014sgldB,teh2016sgld}. The common conclusion of these papers is that the MSE  is of order $\mathcal{O}(c^{-\frac{1}{3}})$ for computational cost of $c$ (as opposed to $\mathcal{O}(c^{-\frac{1}{2}})$ rate of MCMC).

The basic idea of Multilevel Monte Carlo methodology  is to use a cascade of decreasing step-sizes. If those different levels of the algorithm are appropriately coupled, 
one can reduce the computational complexity without a loss of accuracy.

In this paper, we develop a Multilevel SGLD (ML-SGLD) algorithm with computational complexity of $ \mathcal{O}(c^{-\frac{1}{2}})$, hence closing the gap between MCMC and stochastic gradient methods. The underlying idea is based on \cite{SVZ16} and its extensions are:

\begin{itemize}
\item We build an antithetic version of ML-SGLD which removes the logarithmic term present in \cite{SVZ16} and makes the algorithm competitive with MCMC.
\item We consider the scaling of the computational cost as well as the number of data items $N$. By using a Taylor based stochastic gradient, we obtain sub-linear growth of the cost in $N$.
\item By introducing additional time averages, we can speed up the algorithm further.
\end{itemize}
The underlying idea is close in spirit to \cite{Agapiou2014unbiasing}  where expectations of the invariant distribution of an infinite dimensional Markov chain is estimated based on coupling approximations.

This article is organised as follows. In Section 2, we provide a brief description of the SGLD algorithm and the MLMC methodology to extent, which will allow us to sketch in Section 3 how these two ideas can be enmeshed in an efficient way. Next we describe three new variants of the multilevel SGLD with favourable computational complexity properties and study their numerical performance in Section 4. Numerical experiments demonstrate that our algorithm is indeed competitive with MCMC methods which is reflected in the concluding remarks in Section 5.
\section{Preliminaries}
\subsection{Stochastic Gradient Langevin Dynamics}
Let $\theta \in \mathbb{R}^{d}$ be a parameter vector where $\pi(\theta)$ denotes a prior distribution, and $\pi(x|\theta)$ the density of a data item $x$ is parametrised by $\theta$.  By Bayes' rule, the posterior distribution of a set of $N$ data items $X=\{x_{i}\}_{i=1}^{N}$ is given by
$$\pi(\theta|X) \propto \pi(\theta) \prod_{i=1}^{N}\pi(x_{i}|\theta).$$

The following stochastic differential equation (SDE) is ergodic with respect to the posterior  $\pi(\theta|X)$ 
\begin{equation} \label{eq:langevin}
d\theta_{t}=\left( \nabla \log{\pi(\theta_{t})}+\sum_{i=1}^{N} \nabla \log{\pi(x_{i}|\theta_{t})} \right)dt+\sqrt{2}dW_{t}, \quad \theta_0\mathbb{R}^d
\end{equation}
where $W_{t}$ is a $d$-dimensional standard Brownian motion. In other words, the probability distribution of $\theta_t$ converges to $\pi(\theta|X)$ as $t\rightarrow \infty$.
Thus, the simulation of \eqref{eq:langevin} provides an algorithm to sample from $\pi(\theta|X)$. 
Since an explicit solution to \eqref{eq:langevin} is rarely known, we need to discretise it. An application of the Euler scheme yields
\[
\theta_{k+1}=S_{h,\xi_{k}}(\theta_{k}), \quad S_{h,\xi}(\theta):= \theta+h\left( \nabla \log{\pi(\theta)}+\sum_{i=1}^{N} \nabla \log{\pi(x_{i}|\theta)}\right)+\sqrt{2h}\xi
\]
where $\xi_{k}$ is a standard Gaussian random variable on $\mathbb{R}^{d}$. However, this algorithm is computationally expensive since it involves computations on all $N$ items in the dataset. The SGLD algorithm circumvents this problem by replacing the sum of the $N$ likelihood terms by an appropriately constructed sum of $n \ll N$ terms which is given by the following recursion formula
\begin{equation} \label{eq:SGLD}
\theta_{k+1}=S_{h,\tau^{k},\xi_{k}}(\theta_{k}), \quad S_{h,\tau,\xi}(\theta):=\theta+h\left( \nabla \log{\pi(\theta)}+\frac{N}{n}\sum_{i=1}^{n} \nabla \log{\pi(x_{\tau_{i}}|\theta)}\right)+\sqrt{2h}\xi
\end{equation}
with $\xi$ being a standard Gaussian  random variable on $\mathbb{R}^{d}$ and $\tau=(\tau_{1},\cdots, \tau_{s})$ is a random subset of $[N]=\{1,\cdots,N \}$, generated for example by sampling with or without replacement  from $[N]$. Notice that this corresponds to a noisy Euler discretisation. In the original formulation of the SGLD in \cite{welling2011bayesian} decreasing step sizes  $\{h_0\geq h_1 \geq h_2 \geq \ldots \}$ were used in order to obtain an asymptotically unbiased estimator. However, the RMSE is  only of order $\mathcal{O}(c^{-\frac{1}{3}}) $ for the computational cost of $c$ \cite{teh2016sgld}.

\subsection{Multilevel Monte Carlo}
Consider the problem of approximating  $\E[g]$ where $g$ is a random variable. In practically relevant situations, we cannot sample from $g$, but often we can approximate it by another random variable $g^M$ at a certain associated $\mathrm{cost}(g^M)$, which goes to infinity as $M$ increases. At the same time $\lim\limits_{M\to\infty}\E g^M\rightarrow \E g$, so we can have a better approximation, but at a certain cost. The typical biased estimator of  $\E[g]$ then has the form 
\begin{equation} \label{eq:MC}
\hat{g}_{N,M}=\frac{1}{N}\sum_{i=1}^N g^{(M,i)}.
\end{equation}
Consequently, the cost of evaluating the estimator is proportional to $N$ to $\mathrm{cost}(g^M)$. According to the Central Limit theorem, we need to set $N\asymp \epsilon^{-2}\cdot \mathrm{Var}(g^{M})$ to get the standard deviation of the estimator $\hat{g}_{N,M}$ less than $\epsilon$.

Now consider just two approximations $g^M$ and $g^K$, where $K<M$. It is clear, that the cost of one sample for $g^M - g^K$ is roughly proportional to $\mathrm{cost}(g^M)$. We assume that
$V_1=\mathrm{Var} (g^M)\approx \mathrm{Var} (g^K)$ and $V_2=\mathrm{Var} (g^M - g^K)$
where $V_2<V_1$.
Then based on the identity $\E g^M = \E g^K + \E (g^M - g^K)$, we have 
\begin{gather*}
\bar{g}_{N_1,N_2,M,K}=\frac1N_1 \sum_{i=1}^{N_1} g^{(K,i)} + \frac1N_2 \sum_{j=1}^{N_2} \left(g^{(M,j)}-g^{(K,j)}\right).
\end{gather*}
We see that the overall cost of the Monte Carlo estimator $\bar{g}_{N_1,N_2,M,K}$ is proportional to
\begin{gather*}
\mathrm{cost}(\bar{g}_{N_1,N_2,M,K}) = \epsilon^{-2}\cdot \left(\mathrm{cost}(g^K)\cdot V_1 + \mathrm{cost}(g^M)\cdot V_2\right),
\end{gather*}
so implying the condition $$1> \frac{\mathrm{cost}_K}{\mathrm{cost}_M} + \frac{V_2}{V_1},$$ we obtain that
$\mathrm{cost}(\hat{g}_{N,M}) > \mathrm{cost}(\bar{g}_{N_1,N_2,M,K})$. The idea behind this method, which was introduced and analysed in \cite{MR2187308}, lies in sampling $g^M - g^K$ in a way, that $\mathrm{Var} (g^M - g^K)<\mathrm{Var} (g^M)$. This approach has been independently developed by Giles in a seminal work \cite{MG08}, where a MLMC method has been introduced in the setting of stochastic differential equations.

MLMC takes this idea further by using $L\geq 2$ independent clouds of simulations with approximations of a different resolution. This allows the recovery of a complexity $\mathcal{O}(\epsilon^{-2})$ (i.e variance $N^{-1/2}$). The idea of MLMC begins by exploiting the following identity 
\begin{equation} \label{eq:telescoping}
\E[g_L]	= \sum_{l=0}^{L} \E[ g_{l} - g_{l-1} ],  \quad \hbox{with } g_{-1}:=0.
\end{equation}
In our context  $g_l:=g(\theta_T^{M_l})$, $g:\RR^d\rightarrow \RR$,  with  $\{\theta_T^{M_l}\}$, defined in \eqref{eq:SGLD}, 
  $l=0\ldots L$, and $T$ being the final time index in an SGLD sample. We consider a MLMC estimator
\begin{align*} 
   Y = \sum_{l=0}^{L}\left\{ \frac{1}{N_l} \sum_{i=1}^{N_l} \Delta^{(i,l)} \right\}, \quad 	\Delta^{(i,l)}:= g_{l}^{(i)} - g_{l-1}^{(i)}, \quad g_{-1}^{(i)} = 0,
\end{align*}
where $g_{l}^{(i)}= g((\theta_T^{M_l})^{(i)})$ are independent samples at level $l$. The inclusion of the level $l$ in the superscript $(i,l)$ indicates that independent samples are used at each level $l$ and between levels. Thus, these samples can be generated in parallel.

Efficiency of MLMC lies in the coupling of $g_{l}^{(i,l)}$ and $g_{l-1}^{(i,l)}$ that results in small $\mathrm{Var}[\Delta^{(i,l)}]$. In particular, for the SDE in \eqref{eq:langevin}, one can use the same Brownian path to simulate $g_{l}$ and $g_{l-1}$ which, through the strong convergence property of the scheme, yields an estimate for $\mathrm{Var}[\Delta^{(i,l)}]$. More precisely it is shown in Giles \cite{MG08} that under the 
assumptions\footnote{Recall $h_{l}$ denotes the size of the step of the algorithm \eqref{eq:SGLD}.} 
\begin{eqnarray}
\label{ml_ass}
\bigl|\E[g_{l} - g_{l-1} ]| = \mathcal{O}(h_l^{\alpha}),\quad \mathrm{Var}[g_{l} - g_{l-1} ] = \mathcal{O}(h_l^{\beta}), 
\end{eqnarray}
for some \(\alpha\geq 1/2,\) \(\beta>0,\) the expected accuracy under a prescribed computational cost $c$  is proportional to 
\begin{eqnarray*}
\varepsilon\asymp
\begin{cases}
c^{-\frac{1}{2}}, & \beta>\gamma, \\
c^{-\frac{1}{2}}\log^2(c), & \beta=\gamma, \\
c^{-\frac{\alpha}{2\cdot\alpha+\gamma-\beta}}, & 0<\beta <\gamma
\end{cases}
\end{eqnarray*}
where the cost of the algorithm  at each level $l$ is of order $\mathcal{O}(h_l^{-\gamma})$.

The main difficulties  in extending the approach in the context of the SGLD algorithm is a) the fact that  $T \rightarrow \infty$ and therefore all estimates need to hold uniformly in time; b) coupling SGLD dynamics across the different levels in time c) coupling the subsampling across the different levels. All of these problems need serious consideration as naive attempts to deal with them might leave \eqref{eq:telescoping}  unsatisfied, hence violating the core principle of the MLMC methodology.

\global\long\def\ssetfa{\tau^{\mathcal{F},1}}

\global\long\def\ssetfb{\tau^{\mathcal{F},2}}

\global\long\def\ssetf{\tau^{\mathcal{F}}}

\global\long\def\ssetc{\tau^{\mathcal{C}}}
 \global\long\def\b{c}

\global\long\def\iid{\overset{\text{i.i.d.}}{\sim}}

\section{Stochastic Gradient based on Multi-level Monte Carlo}

In the following we present a strategy how the two main ideas discussed
above can be combined in order to obtain the new variants of the SGLD
method. In particular, we are interested in coupling the dicretisations
of \eqref{eq:langevin} based on the step size $h_{l}$ with $h_{l}=h_{0}2^{-l}$.
Because we are interested in computing expectations with respect to
the invariant measure $\pi(\theta|X)$, we also increase the time
endpoint $T_{l}\uparrow\infty$ which is chosen such that $T_{l}/h_{0}\in\mathbb{N}$.
Thus, $s_{l}=T_{l}/h_{l}\in\mathbb{N}$.

We introduce the notation 
\[
S_{h,\tau_{1:s_{l}},\xi_{1:s_{l}}}(\theta_{0})=S_{h,\tau_{s_{l}},\xi_{s_{l}}}\left(S_{h,\tau_{s_{l}},\xi_{s_{l}}}\left(\dots S_{h,\tau_{1},\xi_{1}}(\theta_{0})\right)\right)
\]
where $\xi$ denotes the Gaussian noise and $\tau$ the index of the
batch data. We would like to exploit the following telescopic sum
\[
\mathbb{E}g\left(S_{h,\tau_{1:\theta_{0}},\xi_{1:s_{0}}}(\theta_{0})\right)+\sum_{l}\mathbb{E}g\left(S_{h_{l},\tau_{1:s_{l}},\xi_{1:s_{l}}}(\theta_{0})\right)-\mathbb{E}g\left(S_{h_{l-1},\tau_{1:s_{l-1}},\xi_{1:s_{l-1}}}(\theta_{0})\right).
\]

We have the additional difficulty of different $h_l$  and $h_{l-1}$ stepsizes and simulation time $T_l$.
First, the fine path is
initially evolving uncoupled for $\frac{T_l -T_{l-1}}{h_{l}}$ time steps.
The coupling arises by evolving both fine and coarse paths jointly,
over a time interval of length $T_{l}-T_{l-1}$, by doing two steps
for the finer level denoted by $\theta^{(f,i)}$ (with the time step
$h_{i}$) and one on the coarser level denoted by $\theta^{(c,l)}$
(with the time step $h_{l-1}$) using the discretisation of the averaged
Gaussian input for the coarse step. 

\begin{figure}[!h]
\centering \includegraphics[scale=0.33]{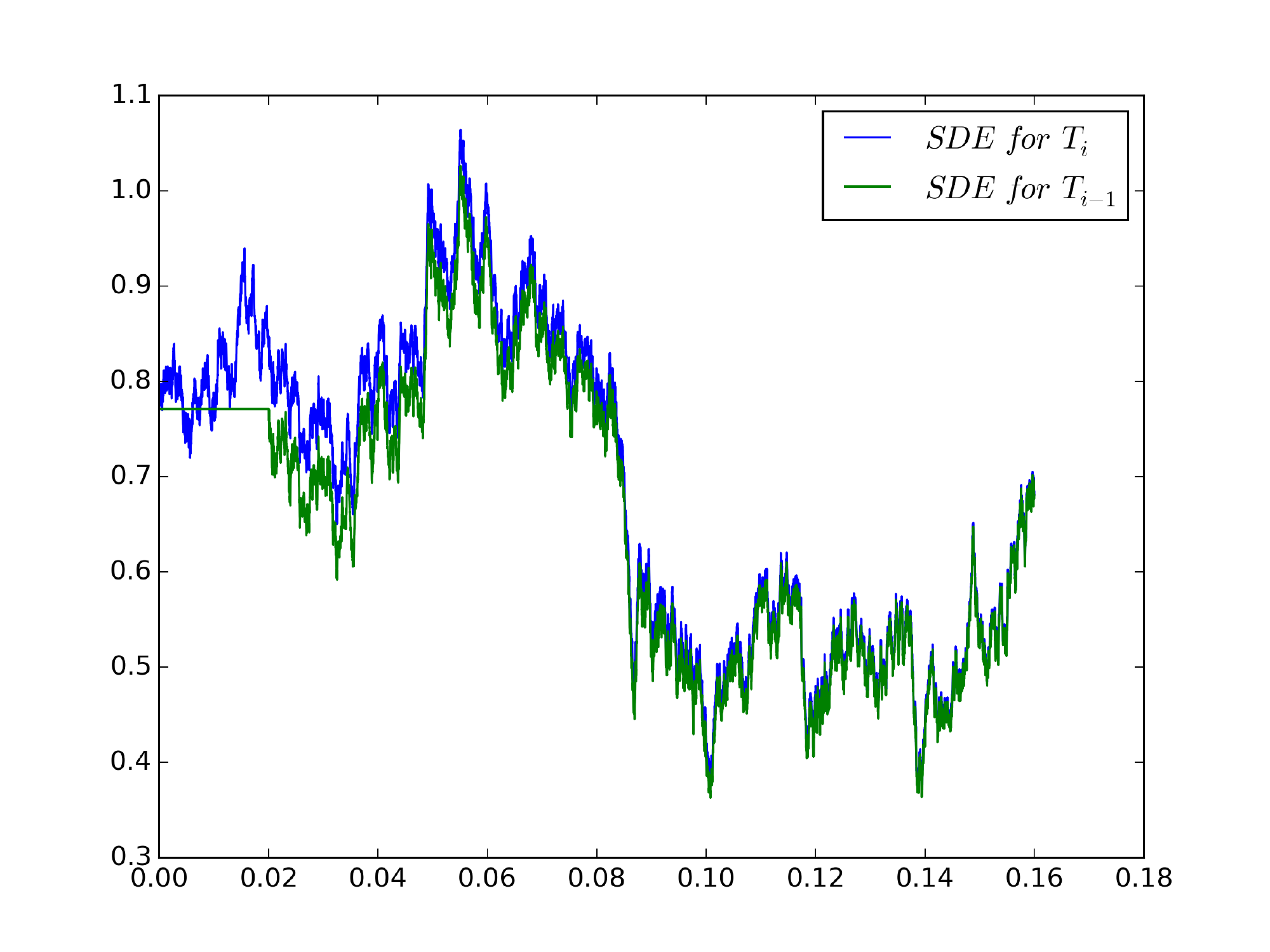} \caption{Behaviour of numerical paths of \eqref{eq:langevin}
when the appropriate coupling is used\label{fig:fig1}}
\end{figure}

This coupling makes use of the underlying contraction (Equation \eqref{eq:contraction}) as illustrated in 
Figure \ref{fig:fig1}. The property that we use is that solutions to \eqref{eq:langevin} started from two different initial conditions $\theta^1_0$ and $\theta^2$ with the same driving noise  satisfy
\begin{equation} \label{eq:contraction}
\mathbb{E} | \theta^1_t - \theta^2_t  |^2 \leq |\theta^1_0 - \theta^2_0 | e^{-L t}, \quad L>0.
\end{equation}
In \cite{SVZ16,2015arXiv150705021D} it is shown that this holds if the posterior is strongly log-concave and also is satisfied by the numerical discretisation. However, numerically this holds for a much larger class and this can be extended by considering more complicated couplings such as the reflection coupling \cite{eberle2011reflection}. This shifting coupling was introduced in \cite{glynn2014} for coupling Markov chains. In \cite{SVZ16,2015arXiv150705021D} it is shown that
\eqref{eq:contraction} holds if the posterior is strongly log-concave.
This is sufficient but not necessary and holds for a much wider class
of problems \cite{eberle2011reflection}.

This property implies that the variance of 
\[
\Delta^{(i,l)}:=g\left(\theta_{\frac{T_{l-1}}{h_{l-1}}}^{(f,l,i)}\right)-g\left(\theta_{\frac{T_{l-1}}{h_{l-1}}}^{(c,l,i)}\right)
\]
for suitably chosen $T_{l-1}$ would remain small, thus allowing an
application of the MLMC methodology. We will drop $i$ appropriately.

\subsection{Multi-level SGLD}

As common in MLMC we couple fine and coarse paths through the Brownian increments, with a Brownian increment on a coarse path given as a scaled sum of increments on the fine - $\frac{1}{\sqrt{2}}\left(\xi_{k,1}+\xi_{k,2}\right)$, which can be written in our notation as
\begin{equation}
\left(\theta_{k+1}^{(f)},\theta_{k+1}^{(c)}\right)=\left(S_{h_{i},\tau_{k,2}^{(f)},\xi_{k,2}}\circ S_{h_{i},\tau_{k,1}^{(f)},\xi_{k,1}}(\theta_{k}^{(f)}),S_{h_{l-1},\tau_{k,1}^{(c)},\frac{1}{\sqrt{2}}\left(\xi_{k,1}+\xi_{k,2}\right)}(\theta_{k}^{(c)})\right).\label{eq:couplingSGLD}
\end{equation}
One question that naturally occurs now is that if and how should one
choose to couple between the subsampling of the data? In particular,
in order for the telescopic sum to be respected, one needs to have
that the laws of distribution for subsampling the data is the same, namely
\begin{equation}
\mathcal{L}\left(\tau^{(f,1)}\right)=\mathcal{L}\left(\tau^{(f,2)}\right)=\mathcal{L}\left(\tau^{(c)}\right).\label{eq:SGLDTelescopic}
\end{equation}
In order for this condition to hold we first take $s$ independent
samples $\tau^{(f,1)}$ on the first fine-step and another $s$ independent
s-samples $\tau^{(f,2)}$ on the second fine-step. In order to ensure
that Equation \ref{eq:SGLDTelescopic} holds, we create $\tau^{(c)}$ by drawing
$s$ samples without replacement from $\left(\tau^{(f,1)},\tau^{(f,2)}\right)$.
Other strategies are also possible and we refer the reader to \cite{SVZ16}.

\begin{algorithm}[h]
\begin{enumerate}
\item The initial steps are characterised by $\hat{s}_{l}=\frac{T_{l}-T_{l-1}}{h_{l}}$ 
\item set $\theta_{0}^{(c,l)}=\theta_{0}$ and $\theta_{0}^{(f,l)}=S_{h_{l},\tau_{-\hat{s}_{l}:-1},\xi_{-\hat{s}_{l}:-1}}$,
then simulate $(\theta_{\cdot}^{(f,l)},\theta_{\cdot}^{(c,l)})$ jointly
according to 
\begin{equation}
\left(\theta_{k+1}^{(f,l)},\theta_{k+1}^{(c,l)}\right)=\left(S_{h_{i},\tau_{k,2}^{(f)},\xi_{k,2}}\circ S_{h_{i},\tau_{k,1}^{(f)},\xi_{k,1}}(\theta_{k}^{(f,l)}),S_{h_{l-1},\tau_{k,1}^{(c)}\frac{1}{\sqrt{2}}\left(\xi_{k,1}+\xi_{k,2}\right)}(\theta_{k}^{(c,l)})\right).\label{eq:langevincoupling}
\end{equation}
and set 
\[
\Delta^{(i,l)}:=g\left(\theta_{s_{l-1}-k}^{(f,l)}\right)-g\left(\theta_{s_{l-1}-k}^{(c,l)}\right).
\]


\end{enumerate}
\protect\caption{\label{alg:CouplingLangevinDiscretisation} ML-SGLD for $t_{i}\uparrow\infty$}
\end{algorithm}

\subsection{Antithetic Multi-level SGLD}

Here we present the most promising variant of coupling on subsampling:
Algorithm \ref{alg:Anithetic}  for $t_{i}\uparrow\infty$.
Building on the ideas developed in \cite{MR3211005} (see also \cite{MR3349310})
we propose Antithetic Multi-level SGLD which achieves an MSE of order 
complexity $\mathcal{O}(c^{-\frac{1}{2}})$ for prescribed computational cost  (and therefore allows for
MLMC with random truncation see \cite{Agapiou2014unbiasing}).

\begin{algorithm}[h]
\begin{enumerate}
\item The initial steps are characterised by $\hat{s}_{l}=\frac{T_{l}-T_{l-1}}{h_{l}}$ 
\item set $\theta_{0}^{(c+,l)}=\theta_{0}^{(c-,l)}=\theta_{0}$ and $\theta_{0}^{(f,l)}=S_{h_{l},\tau_{-\hat{s}_{l}:-1}},\xi_{-\hat{s}_{l}:-1}$,
then simulate $(\theta_{\cdot}^{(f,l)},\theta_{\cdot}^{(c,l)})$ jointly
according to

\begin{align}
\begin{aligned}\label{eq:Antitheitc}\theta_{k+1}^{(f,i)} & =S_{h_{l},\tau_{k,2},\xi_{k,2}^{(f)}}\circ S_{h_{i},\tau_{k,1}^{(f)},\xi_{k,1}}(\theta_{k}^{(f,l)})\\
\theta_{k+1}^{(c+,l)} & =,S_{h_{l-1},\tau_{k,1}^{(f,1)}\frac{1}{\sqrt{2}}\left(\xi_{k,1}+\xi_{k,2}\right)}(\theta_{k}^{(c,l)})\\
\theta_{k+1}^{(c-,l)} & =,S_{h_{l-1},\tau_{k,1}^{(f,2)}\frac{1}{\sqrt{2}}\left(\xi_{k,1}+\xi_{k,2}\right)}(\theta_{k}^{(c,l)})
\end{aligned}
\end{align}

\item set 
\[
\Delta^{(i,l)}:=g\left(\theta_{\frac{t_{l-1}}{h_{l-1}}-k}^{(f,l)}\right)-\frac{1}{2}\left(g\left(\theta_{\frac{t_{l-1}}{h_{l-1}}-k}^{(c+,l)}\right)+g\left(\theta_{\frac{t_{l-1}}{h_{l-1}}-k}^{(c-,l)}\right)\right).
\]

\end{enumerate}
\protect\caption{\label{alg:Anithetic} Antithetic ML-SGLD for $t_{l}\uparrow\infty$}
\end{algorithm}

\subsection{Averaging the Path}\label{sec:averaging}

Compared to MCMC these algorithms seem wasteful because only the last
step of a long simulation is saved. The numerical performance can
be improved by instead averaging of parts of the trajectory as follows
\[
\Delta_{\text{averaged}}^{(i,l)}:=\frac{1}{p_{l}}\sum_{k=0}^{p_{l}}g\left(\theta_{\frac{t_{l-1}}{h_{l-1}}-k}^{(f,l)}\right)-\frac{1}{p_{l-1}}\sum_{k=0}^{p_{l-1}}g\left(\theta_{\frac{t_{l-1}}{h_{l-1}}-k}^{(c,l)}\right),
\]
and this also applies appropriately to the antithetic version.

\subsection{Taylor based Stochastic Gradient}\label{sec:taylor}

The idea of Taylor based stochastic gradient is to use subampling
on the remainder of a Taylor approximation 
\begin{align}
 & \sum_{i=1}^{N}\nabla\log p\left(x_{i}\vert\theta\right)\nonumber \\
 & =\sum_{i=1}^{N}\nabla\log p\left(x_{i}\vert\theta_{0}\right)+\sum_{i=1}^{N}\nabla^{2}\log p(x_{i}\vert\theta_{0})\left(\theta-\theta_{0}\right)\nonumber \\
 & +\sum_{i=1}^{N}\left(\nabla\log p\left(x_{i}\vert\theta\right)-\left(\nabla\log p\left(x_{i}\vert\theta_{0}\right)+\nabla^{2}\log p(x_{i}\vert\theta_{0})\left(\theta-\theta_{0}\right)\right)\right)\nonumber \\
 & \approx\sum_{i=1}^{N}\nabla\log p\left(x_{i}\vert\theta_{0}\right)+\left(\sum_{i=1}^{N}\nabla^{2}\log p(x_{i}\vert\theta_{0})\right)\left(\theta-\theta_{0}\right)\label{eq:Taylor}\\
 & +\frac{N}{n}\sum_{i=1}^{n}\left(\nabla\log p\left(x_{\tau_{i}}\vert\theta\right)-\left(\nabla\log p\left(x_{\tau_{i}}\vert\theta_{0}\right)+\nabla^{2}\log p(x_{\tau_{i}}\vert\theta_{0})\left(\theta-\theta_{0}\right)\right)\right).\nonumber 
\end{align}
We expect that the Taylor based stochastic gradient to have small
variance for $\theta-\theta_{0}$ small. The idea of subsampling the
remainder originally was introduced in \cite{sinan2015}. By interopolating
between the Taylor based stochastic gradient and the standard stochastic
gradient we have the best of both worlds.

\section{Experiments}
\label{sec:exp} \global\long\def\data{y}
 \global\long\def\logit{f}
 \global\long\def\cova{\iota}

We use Bayesian logistic regression as testbed for our newly proposed methodology and perform a simulation study. The 
data $\data_{i}\in\{-1,1\}$ is modelled by 
\begin{equation}
p(\data_{i}\vert\iota_{i},x)=\logit(y_{i}x^{t}\cova_{i})\label{eq.logistic}
\end{equation}
where $\logit(z)=\frac{1}{1+\exp(-z)}\in[0,1]$ and $\cova_{i}\in\mathbb{R}^{d}$
are fixed covariates. We put a Gaussian prior $\mathcal{N}(0,C_{0})$
on $x$, for simplicity we use $C_{0}=I$ subsequently. By Bayes'
rule the posterior $\pi$ satisfies 
\[
\pi(x)\propto\exp\left(-\frac{1}{2}\left\Vert x\right\Vert _{C_{0}}^{2}\right)\prod_{i=1}^{N}\logit(y_{i}x^{T}\cova_{i}).
\]
We consider $d=3$ and $N\in\{100,316,1000,3162,10000\}$ data points
and choose the covariate to be 
\[
\cova=\left(\begin{array}{ccc}
\cova_{1,1} & \cova_{1,2} & 1\\
\cova_{2,1} & \cova_{2,2} & 1\\
\vdots & \vdots & \vdots\\
\cova_{N,1} & \cova_{N,2} & 1
\end{array}\right)
\]
for a fixed sample of $\cova_{i,j}\overset{\text{i.i.d.}}{\sim}\mathcal{N}\left(0,1\right)$
for $i=1,\dots N$ and we take $n=\left\lceil N^{\frac13}\right\rceil $.

It is reasonable to start the path of the individual SGLD
trajectories at a mode of the target distribution. This means that we set the
$x_{0}$ to be the map estimator 
\[
x_{0}=\text{argmax}\:\exp\left(-\frac{1}{2}\left\Vert x\right\Vert _{C_{0}}^{2}\right)\prod_{i=1}^{N}\logit(y_{i}x^{T}\cova_{i})
\]
which is approximated using the Newton-Raphson method. In the following we disregard the cost for the preliminary computations which could be reduced using state of the art optimisation and evaluating the Hessian in parallel. In the 
following we use MCMC and the newly developed MLSGLD to estimate the averaged squared distance from the map estimator under the posterior $\int_{\mathbb{R}^3} \|\theta-\theta_0\|^2 \pi(x)dx$ i.e. set 
\begin{equation}\label{eq:g}
g(\theta)=\|\theta-\theta_0\|^2.
\end{equation}
Notice that by posterior consistency properties we expect this quantity to be have like $\frac{1}{N}$ which is why we will consider relative MSE.

\subsection{Illustration of Coupling standard, antithetic and with Taylor}
\begin{figure}
\subfloat[Coupled levels]{\label{fig:1c}\includegraphics[height=4.5cm, width=0.33\textwidth]{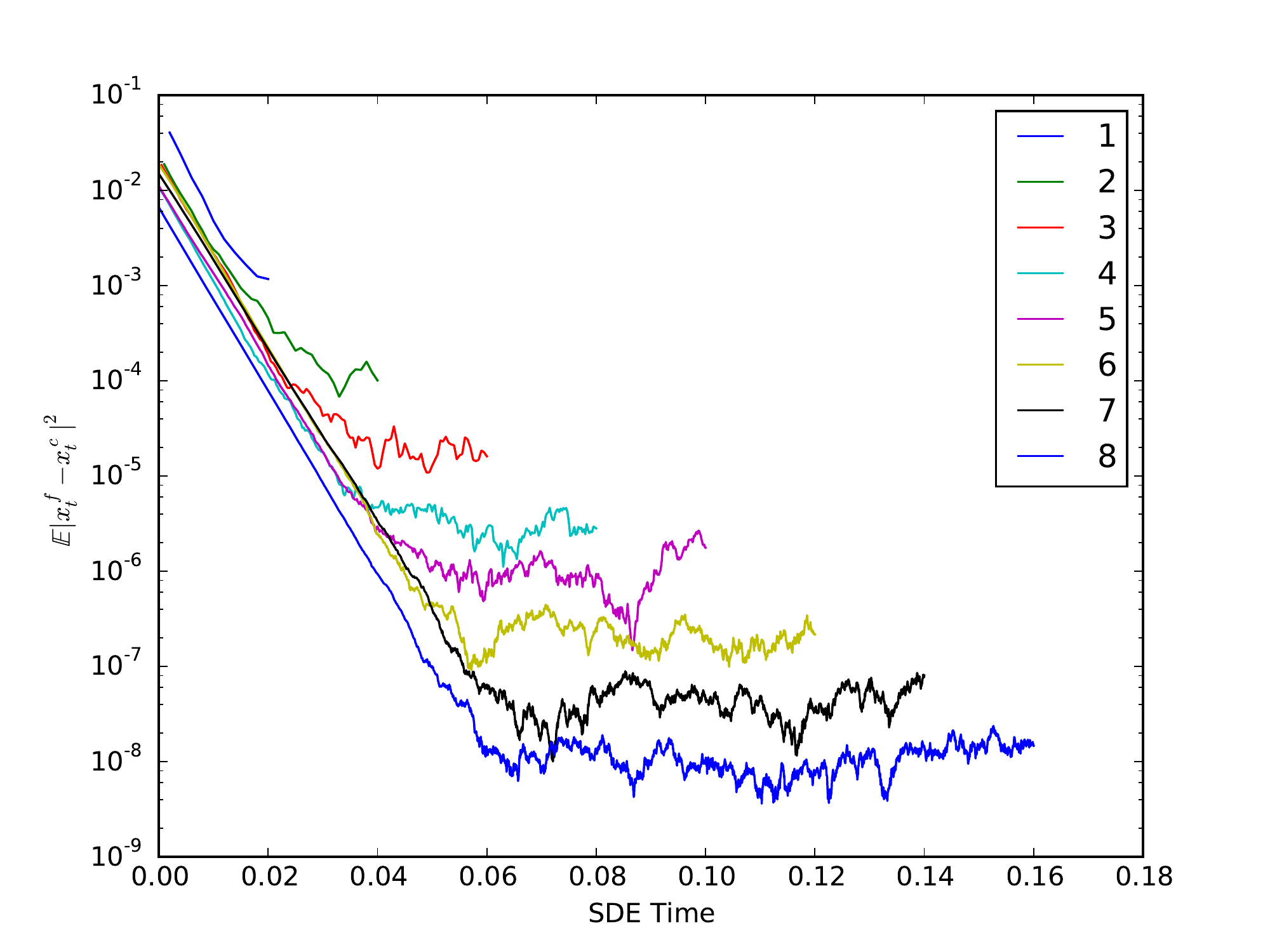}}\subfloat[Variance decay]{\label{fig:1b}\includegraphics[height=4.5cm, width=0.33\textwidth]{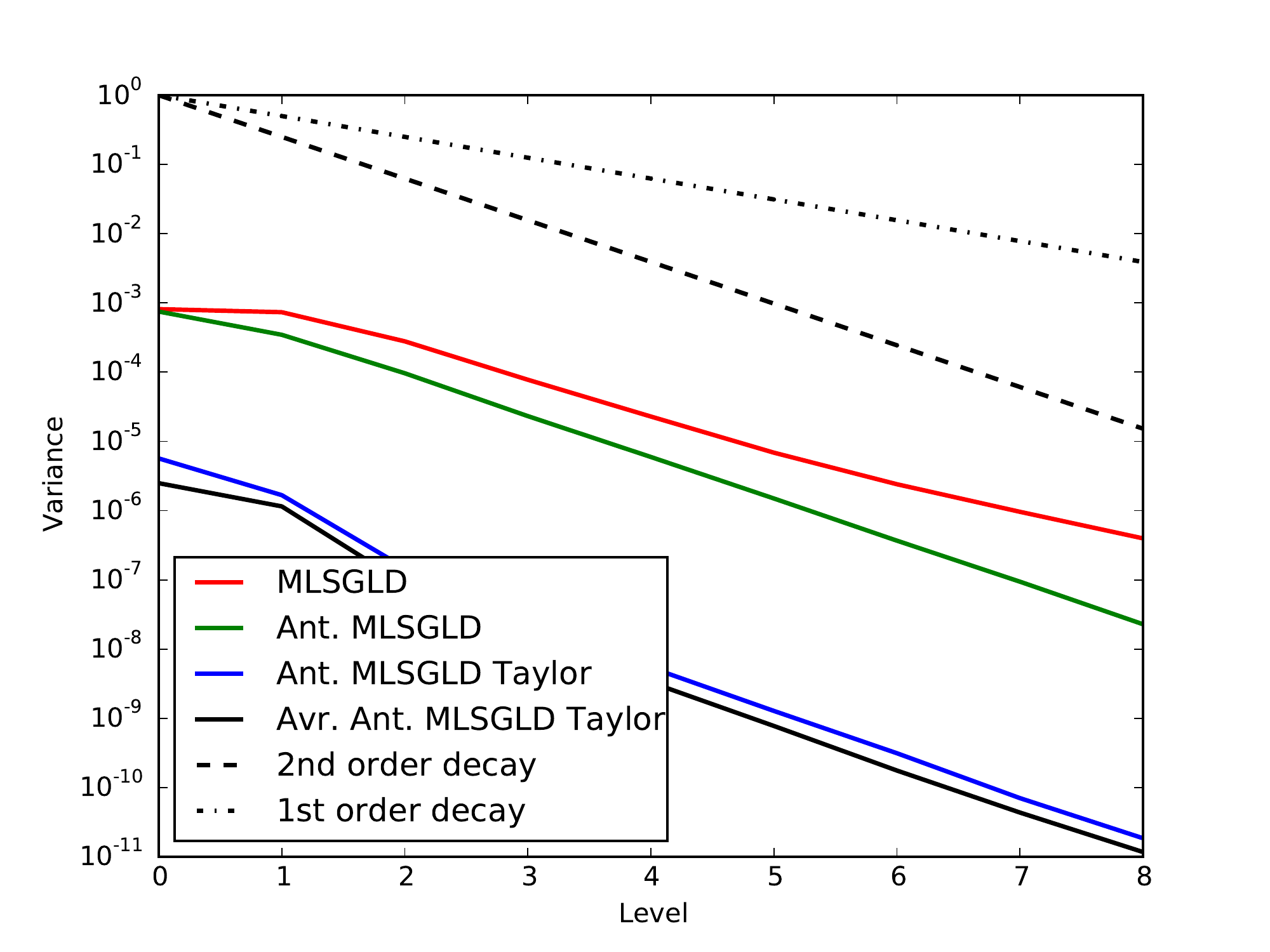}
}\subfloat[Mean decay]{\label{fig:1a}\includegraphics[height=4.5cm, width=0.33\textwidth]{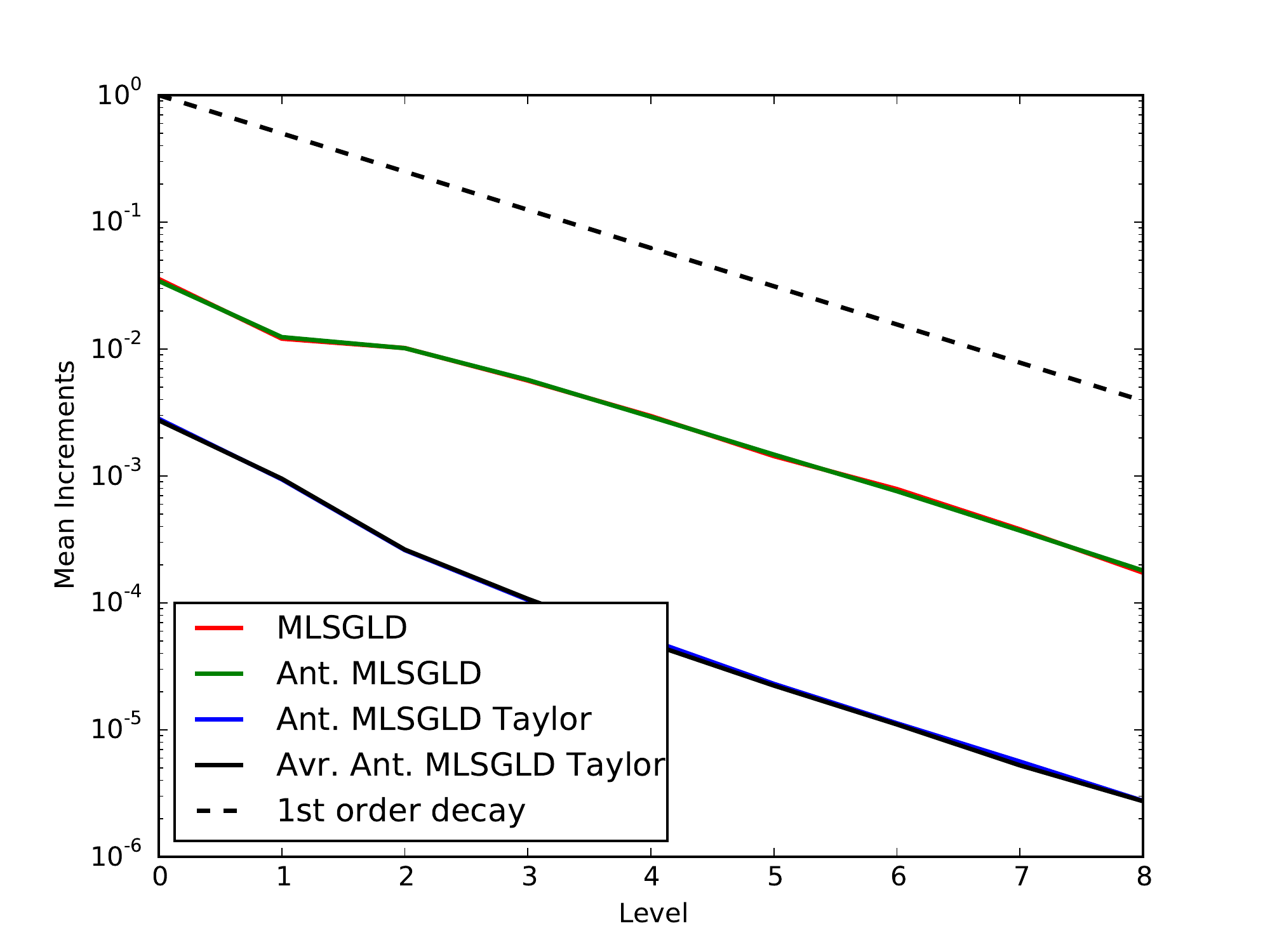}
}
\caption{Coupled paths at differen levels, variance and mean decays with respect to the levels.}
\label{fig:decay}
\end{figure}

We choose $T_{l}= m(l+1)h_0$, $h_l=2^{-l}$ and leave $m\in \mathbb{N}$ as a tuning parameter. The crucial ingredient here is that in expectation  the coarse and fine paths get closer exponentially initially and then asymptote, with the asymptote decaying as the step size decays. This illustrated on Figure \ref{fig:1c}. As any MLMC algorithm performance is effected by the order $\beta$ of the variance $\mathrm{Var}\Delta^{(i,l)}\preceq h_l^\beta$, the parameters $m$ and $h_0$ should be chosen such that the difference between pathes reaches the asymptote, but preferrably does not spent to much time in it, as this increases the computational cost of sampling those paths. In our experiments we set $m=5$ and $h_0 = 1/N$ and on Figure \ref{fig:1a} we see, that Algorithm \ref{alg:Anithetic} provides better coupling with variance decay of order $2$, which is significantly better than the first order variance decay, given by Algorithm \ref{alg:CouplingLangevinDiscretisation}. Combining Algorithm \ref{alg:Anithetic} with Taylor based extension from Section \ref{sec:taylor} and path averaging with $p_l = s_l/2$ from Section \ref{sec:averaging} gives additional decrease for the variance without affecting the rate $2$. The faster variance decay leads to lower overall complexity, as the number of samples at each level is proportional to the variance at that level. The Taylor Mean decay rates are of the same order, which can be seen on Figure \ref{fig:1b}, but once again Algorithm \ref{alg:Anithetic} combined with Taylor and path averaging is more preferable, as the multiplicative constant is lower, than in Algorithm \ref{alg:CouplingLangevinDiscretisation}.

Numerical evidence, presented here, leads to the conclusion, that Antithetic MLSGLD with Taylor along with Antithetic MLSGLD with Taylor and Averaging are the best competitors to MCMC algorithm, so we proceed to comparison of those algorithms.
\subsection{Comparison with MCMC}
\begin{figure}
\subfloat[Comparison of Mean Squared Relative Errors ]{\label{fig:2a}\includegraphics[height=4.1cm]{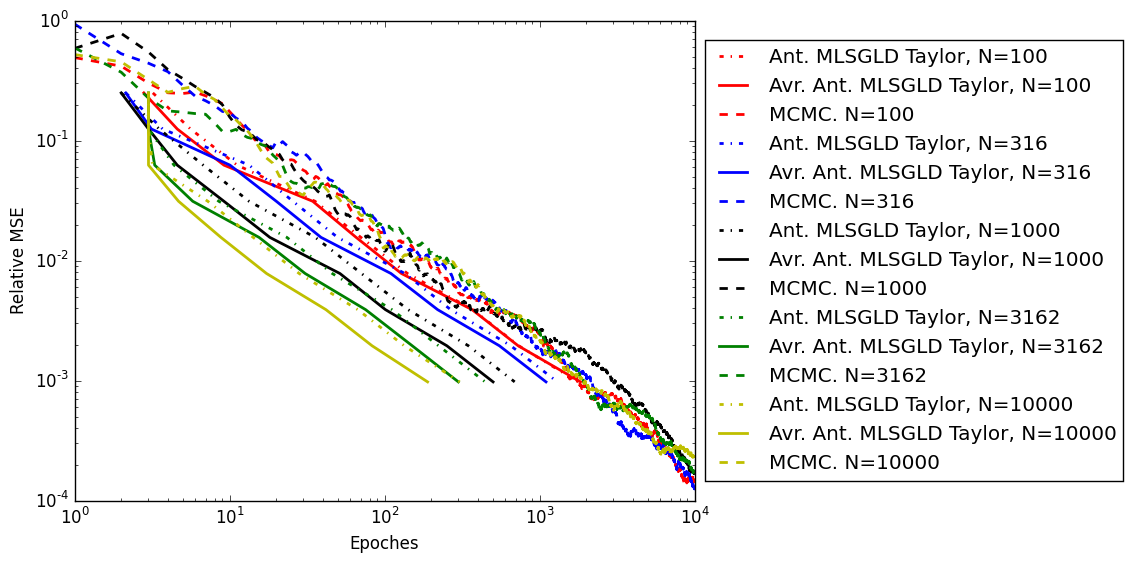}
}\subfloat[Increase in Complexity]{\label{fig:2b}\includegraphics[height=4.1cm]{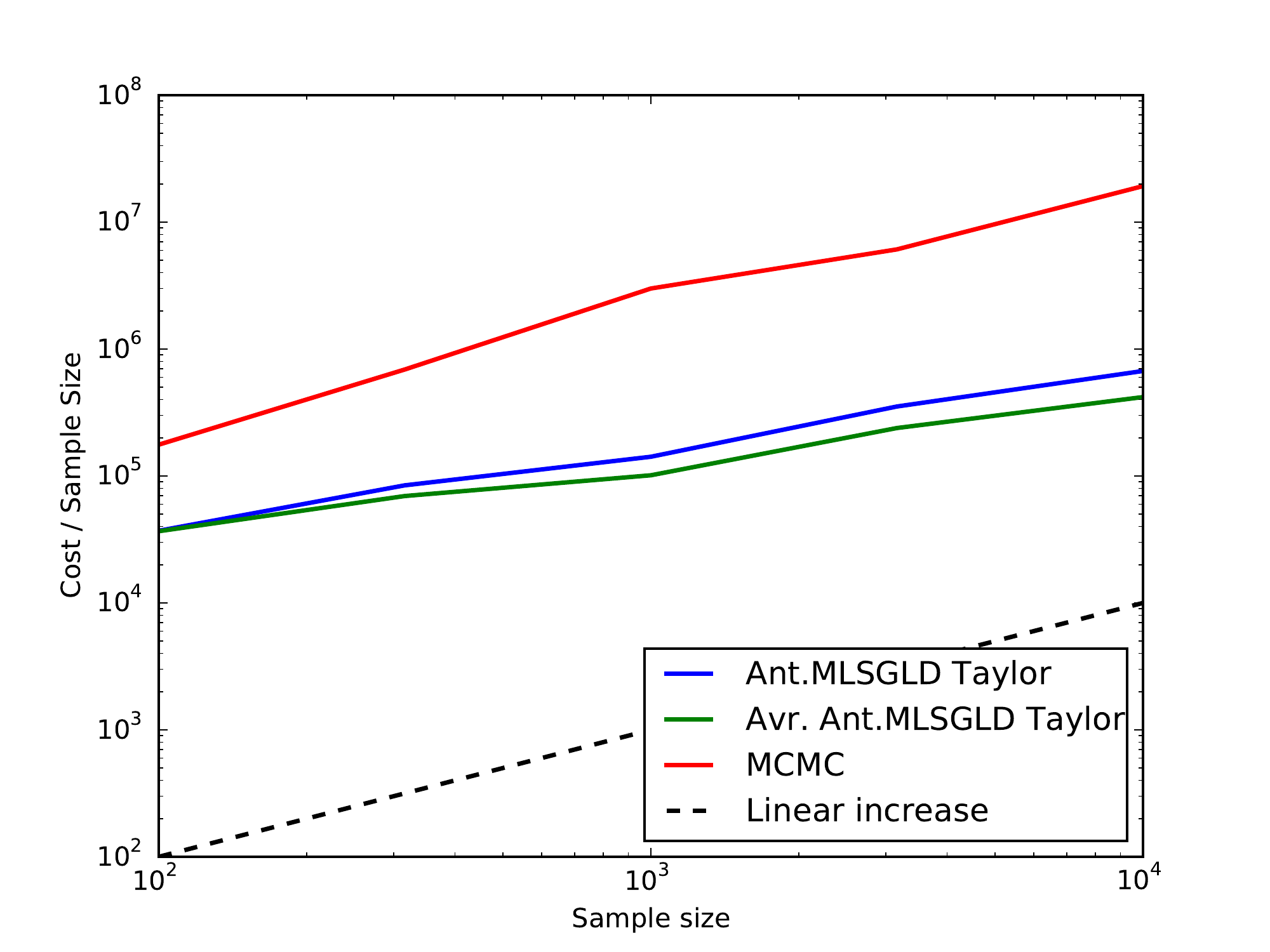}
}
\caption{\label{fig:MCMC}. Scalability of the algorithms and achieved Relative MSE for for differen datasets.}
\end{figure}

We choose Metropolis-Adjusted Langevin (MALA, see \cite{roberts1996exponential}) as a competitor because it is based on one Euler step of the  Langevin SDE, but adds a Metropolis accept-reject step in order to preserve the correct invariant measure (removing the requirement to decrease step size for better accuracy). We take cost as the number of evaluation of data items, which is typically measured in epochs. One epoch corresponds to one effective iteration through the full data set. Heuristically, for this log-concave problem we expect the convergence rate to be independent of $N$, so the only cost increase is due to evaluating posterior density and evaluating $\nabla \log \pi (X|\theta) $. This agrees with the findings in Figure \ref{fig:2a}, where the MCMC lines are almost on top of each other thus yielding the same relative MSE for the same number of epochs for different dataset sizes. As $N$ increases the cost per epoch increases proportional to $N$. We run the MALA for $10^4$ steps with $10^3$ steps of burning and optimal acceptance rate $0.574$ for 50 times and then average. The various MLSGLD algorithms are ran for 50 times to achieve relative accuracies $2^{-k/2},\ k=2,\ldots,10$. This is yet another advantage of MLMC paradigm, which allows us to control numerically the mean increments and variance at all the levels, thus stopping the algorithm, when it has converged numerically.
The most important comparison is presented on Figure \ref{fig:2b}, where we compare the increase of the complexity to achieve relative accuracy of $2^{-5}$ with respect to the dataset size. We observe the sublinear growth of cost w.r.t dataset size for Antithetic MLSGLD with Taylor and Antithetic MLSGLD with Taylor and averaging, with the later having a slightly better behaviour than the first one.
\section{Conclusion}

We develop a Multilevel SGLD algorithm with computational complexity of $ \mathcal{O}(c^{-\frac{1}{2}})$, hence closing the gap between MCMC and stochastic gradient methods. Moreover, this algorithm scales sublinearly with respect to the dataset size and allows natural parallelization, due to the typical properties of Monte Carlo sampling. The benefits of parallelization are to be studied later along with further numerical investigations for adaptive choices of parameters in the algorithm. In our further studies we also plan to quantify analytically the gains, given by MLSGLD algorithm and extend its applicability to a larger class of models.
\bibliographystyle{abbrv} 
\bibliography{refs}
\end{document}